\documentclass{esannV2}
\usepackage[dvips]{graphicx}
\usepackage[latin1]{inputenc}
\usepackage{amssymb,amsmath,array}
\usepackage{titlesec}
\usepackage{eso-pic}
\usepackage{url}
\titlespacing*{\section}{0pt}{1.0ex}{0.1ex}
\titlespacing*{\subsection}{0pt}{1ex}{0pt}
\begin{document}
\AddToShipoutPictureFG*{%
  \AtPageUpperLeft{%
    \put(80,-30){%
      \makebox[\paperwidth][c]{%
        \footnotesize Submitted to ESANN 2027.
      }%
    }%
  }%
}
\title{Effectiveness of Synthetic Defects for Anomaly Detection in Industrial Small Load Carriers}

\author{
Vincent Wahyudi$^1$, Kuniko Paxton$^2$, Amila Akagic\textsuperscript{3,4},
\\ Martin Storath$^1$, and Dhavalkumar Thakker$^2$
%
%
\vspace{.3cm}\\
%
1- Technische Hochschule W\"urzburg-Schweinfurt, 
\\ Ignaz-Sch\"on-Stra{\ss}e 11, 97421 Schweinfurt - Germany
\vspace{.1cm}\\
2- University of Hull, Cottingham Road, Hull, HU6 7RX - United Kingdom\\
%
3- University of Sarajevo, 
\\ Obala Kulina bana 7/II, 71000, Sarajevo - Bosnia and Herzegovina\\
4- College of Computing and IT, University of Doha for Science and Technology, 
\\ Al Tarafa, Jelaiah Street Duhail North, Doha - Qatar
}


\maketitle

\begin{abstract}
Recent industrial anomaly detection methods use procedural noise to generate synthetic training defects. Their effectiveness in real-world settings and the impact of different noise generation methods have not been sufficiently investigated. We evaluate the effectiveness of training using synthetic defects for both supervised and unsupervised methods, the impact of different noise generation methods, and the differences in detection performance between benchmark datasets and a real-world small load carrier (SLC) dataset. Trained on synthetic defects only, a supervised classifier reaches 0.65 AUROC on real SLC defects, while three unsupervised methods using the same kind of noise (Perlin) reach 0.87--0.90. Except for SuperSimpleNet, which fails with the foreground-confined fractal Brownian motion (fBm) recipe, models trained with the custom recipe or with nine other noise types do not differ reliably from those trained with Perlin noise. Finally, results with synthetic defects on benchmark datasets do not carry over to the SLC dataset, possibly because the synthetic defects differ too much from the real ones (a larger angle in feature space). These results call for caution when training models on synthetic defect images.
\end{abstract}

\section{Introduction}
\label{sec:introduction}
Quality control (QC) of reusable small load carriers (SLCs) relies on manual visual inspections, which are prone to error from fatigue and inconsistent judgement criteria. Wahyudi et al. \cite{wahyudi2025computer} explored automated QC. Since real anomalous SLC images are scarce, they evaluated several unsupervised anomaly detection methods on images acquired in a real-world environment.

In recent years, researchers have compensated for the shortage of real anomalous samples by generating synthetic defects from OK images \cite{zhang2023destseg, chen2024unified}. For example, CutPaste \cite{li2021cutpaste} generates pseudo-anomalies by cutting and pasting image regions, while DRAEM \cite{zavrtanik2021draem} generates synthetic defects by filling Perlin noise \cite{perlin2002improving} masks with external textures. However, it is unclear whether the choice of noise generator matters, and whether it matters if synthetic defects serve as labels (supervised) or only as a training signal (unsupervised).

Anomaly detection methods using synthetic defects are typically evaluated on benchmark datasets such as MVTec AD \cite{bergmann2019mvtec}. However, \cite{baitieva2025beyond,pemula2025robust} reported a gap between benchmark performance and real-world industrial applications. Several factors set the SLC dataset apart. As SLCs are reused, wear and discolouration occur even on OK ones. Diverse defect types and a changing environment during operation can also widen the gap. This study investigates the effectiveness of synthetic defects generated from ten noise types and blending methods, as labels for a supervised classifier and as a training signal for unsupervised methods, on benchmark datasets and the SLC dataset. 



\section{Method and Experimental Setup}
\subsection{Datasets}
The real-world dataset we use is a subset of the public SLC dataset \cite{wahyudi2025computer}. It contains four SLC classes (18, 289, 120 and 194), and each SLC has up to five views, resized to 256 $\times$ 256. Labels are given at image and sample level, without pixel masks. In total, the subset has 3085 images (1929 OK and 1156 defective). As a benchmark, we use the 15 MVTec AD \cite{bergmann2019mvtec} categories (5 textures, 10 objects) for the supervised classifier. 

\subsection{Models}
We tested three unsupervised methods: DRAEM \cite{zavrtanik2021draem}, SuperSimpleNet (SSN) \cite{rolih2024supersimplenet} and DSR \cite{zavrtanik2022dsr}. Each model injects the synthetic defects at a different stage. DRAEM thresholds the noise (Perlin by default) into a pixel mask and blends an external texture into the image inside it. SSN uses the noise as a mask in feature space and adds Gaussian noise to the features inside it. DSR uses the noise as a mask in its quantised feature space and replaces the feature codes inside it. In the supervised setting, the synthetic defects serve as labelled defects for a linear classifier (logistic regression), trained per class on frozen \mbox{WideResNet-50-2} features.

\begin{figure}[h!]
\centering
\includegraphics[scale=0.7]{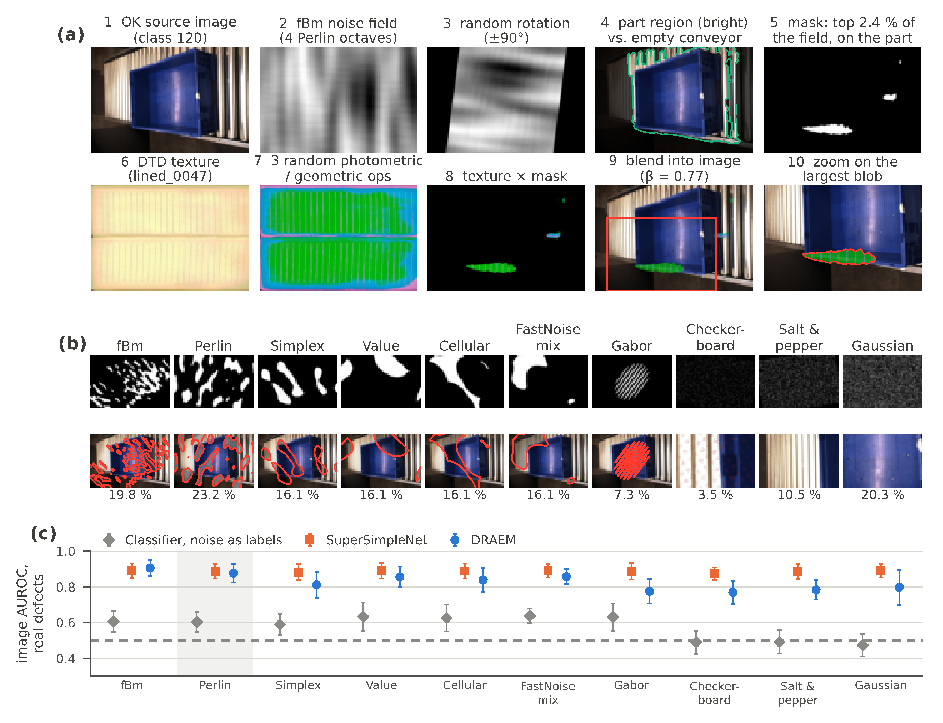}
\caption{How a synthetic defect is made. (a) Our main recipe \texttt{fg\_dtd\_matched} step by step on one SLC image: an fBm noise field is thresholded into a mask that lies mainly on the SLC (fg) and covers 1--3\% of the image, matched to the measured size of real defects. The mask is filled with an altered texture from the Describable Textures Dataset (DTD) \cite{cimpoi2014describing} and blended in with opacity 0.1--0.8. (b) The 10 noise kinds tested under the default recipe (mask 5--25\% of the image, anywhere, no DTD): mask (top) and resulting image with the mask outlined in red (bottom). (c) Image AUROC on real defects for each noise kind on the 4 SLC classes. The noise kind barely changes SSN. DRAEM varies more, but no noise kind differs reliably from Perlin.}
\label{Fig:generation}
\end{figure}

\subsection{Anomaly Image Generation}
Figure \ref{Fig:generation}a shows step by step how our recipe generates defects. Steps 2, 4, and 5 are our contribution. Steps 3 and 6 to 9 follow DRAEM's setup. The mask is placed mainly on the SLC instead of the background, and its size matches that of real defects as a share of the image. The mask is filled with a DTD \cite{cimpoi2014describing} texture and blended in with opacity 0.1--0.8. The models use the generated images and masks at different stages. We compare synthetic and real defects in the classifier's (standardised) features by the
angle $\theta=\arccos\frac{d_\mathrm{syn}\cdot d_\mathrm{real}}{\lVert d_\mathrm{syn}\rVert\lVert d_\mathrm{real}\rVert}$,
where $d_\mathrm{real}$ is the mean real-defect minus mean OK features (per camera, weighted by defect count) and $d_\mathrm{syn}$ the mean difference between each synthetic image and its OK source. An angle of 0\textdegree{} means the same change, 90\textdegree{} an unrelated one. 

Depending on the noise generation method, characteristics such as spatial correlation, frequency response, directionality, connectivity, periodicity, and scale vary \cite{lagae2010survey}. We use ten types of noise and artificial patterns: Perlin noise, Simplex noise \cite{perlin2002noisehardware}, Value noise \cite{572337}, fractal Brownian motion (fBm) \cite{gonzalezvivo2015bookofshaders}, Cellular noise \cite{worley1996cellular}, Gabor noise \cite{lagae2010improving}, Gaussian noise \cite{szeliski2022computer}, salt-and-pepper noise \cite{gonzalez2008digital}, checkerboard patterns, and FastNoise Mix (Perlin/Simplex/Value maps, picked at random). Figure \ref{Fig:generation}b shows the generated masks on the SLC dataset.

\begin{figure}[h!]
\centering
\includegraphics[scale=0.7]{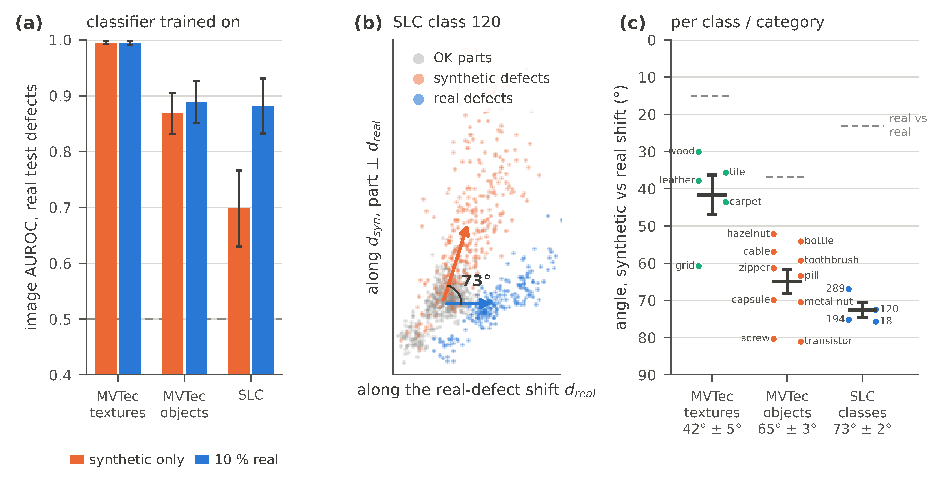}
\caption{(a)~Image AUROC on real test defects for a classifier trained on synthetic defects
only vs.\ on 10\% of the real ones; mean $\pm$ SE over classes. (b)~SLC class~120 in the plane of $d_\mathrm{real}$ and $d_\mathrm{syn}$ (true angle). (c)~Synthetic-vs-real shift angle per class/category; black: group mean $\pm$ SE; dashed: real-vs-real angle between two halves of the real defects (noise floor).}
\label{Fig:angle}
\end{figure}

\begin{figure}[h!]
\centering
\includegraphics[scale=0.75]{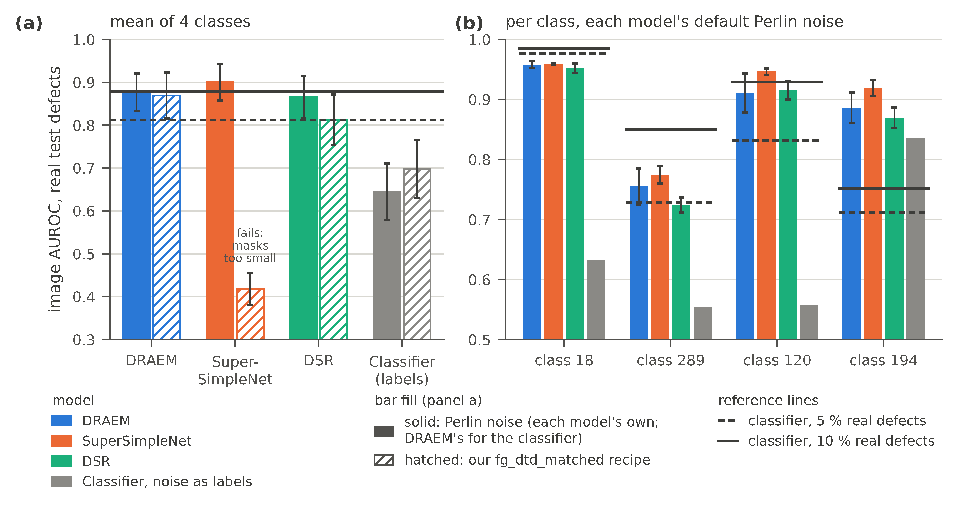}
\caption{(a) Mean of the 4 classes; error bars = standard error over the 4 classes (each class's AUROC is the mean of 3 runs). Solid: each unsupervised model with its default Perlin noise (the generator built into the method), trained on OK images only, and in grey the classifier trained on DRAEM's Perlin noise as labels. Hatched: our recipe \texttt{fg\_dtd\_matched}. Lines: the classifier trained with 5\% (dashed) or 10\% (solid) of the real training defects. (b) The same per class, default Perlin noise only. Same test images for all.}
\label{Fig:unsupervised}
\end{figure}

\section{Results and Discussion}
Figure \ref{Fig:generation}c compares the 10 noise kinds. SSN does not depend on the noise kind (0.87--0.89, all within $\pm$0.014 of Perlin). In DRAEM, fBm (0.91) and Perlin (0.88) score highest, but no kind differs reliably from Perlin. The classifier trained on synthetic defects stays at 0.47--0.64, close to chance (0.5) at the low end.
Figure \ref{Fig:angle}a shows the classifier trained on synthetic or real defects. Trained on real OK images and synthetic defects only, and tested on real defects, it reaches 0.70 AUROC; adding 5\%/10\% of real defects increases this to 0.81/0.88. On MVTec AD textures, synthetic only and 10\% real are level (0.995 vs 0.994); on MVTec AD objects, synthetic only is slightly below 10\% real (0.87 vs 0.89). Figure \ref{Fig:angle}b shows that the synthetic defects for SLC class 120 point 73\textdegree{} away from the real defects. In Figure \ref{Fig:angle}c, the dashed line at 23\textdegree{} is the SLC angle between two halves of the real defects. The angle between synthetic and real defects is smaller for MVTec AD textures (42\textdegree) and objects (65\textdegree) than for SLC (73\textdegree), and so is its gap to the real-vs-real line (27\textdegree{} and 28\textdegree{} vs 50\textdegree{}). Across the three groups, a larger angle goes with a lower synthetic-only AUROC.

Figure \ref{Fig:unsupervised}a shows that the unsupervised methods with their default Perlin noise and no real defects reach 0.87--0.90, close to the classifier with 10\% real (0.88). With our recipe, SSN fails (0.42), while DSR and DRAEM do not differ reliably from default Perlin. Figure \ref{Fig:unsupervised}b shows that with 10\% of the real defects the classifier beats the unsupervised methods on classes 18 and 289, is level on class 120 and falls behind on class 194.


\section{Conclusion}
Noise-generated synthetic defects were poor training positives for the supervised classifier on SLCs: training on them reached only 0.70 AUROC, while 10\% of the real defects reached 0.88. On MVTec AD textures, the same classifier reached 0.995 with synthetic only and 0.994 with 10\% real defects. So results with synthetic defects on the benchmark do not carry over to the SLC dataset, possibly because in feature space the synthetic defects lie much further from the real ones on SLC (73\textdegree{}) than on MVTec AD textures (42\textdegree).

Used inside the unsupervised methods (DRAEM, DSR and SSN), defects from their default Perlin noise reached 0.87--0.90 AUROC, well above the supervised classifier trained on DRAEM's Perlin defects (0.65). With our recipe \texttt{fg\_dtd\_matched}, DRAEM and DSR reached 0.87 and 0.81, while SSN failed because its masks were too small. The recipe did not give the unsupervised methods an advantage over default Perlin noise, and none of the other noise kinds differed reliably from Perlin. For these methods, synthetic defects apparently only need to differ from OK samples. Since none of our generators came close to real defects, whether realism would help remains open. Results with synthetic defects on benchmark datasets should therefore be checked on real industrial data, and their value depends on the training method: they work better as a training signal for unsupervised methods than as labels for a supervised classifier.

\section*{Acknowledgment}
The authors thank the EPSRC National Edge AI Hub. Kuniko Paxton appreciates financial support from COST Action CA22143 (STSM for a research visit to the University of Sarajevo to undertake training for this study). The contribution of Vincent Wahyudi to this work is also supported by DIG:IT-KMU (Grant No. 20-3044-2-23), funded by the Bavarian Ministry of Economic Affairs, Regional Development and Energy under the EFRE Bayern 2021--2027 programme and co-financed by the EU. The authors declare no conflicts of interest.


\begin{footnotesize}


\bibliographystyle{unsrt}
\bibliography{reference}

\end{footnotesize}


\end{document}